\documentclass[journal]{IEEEtran}
\usepackage{amsmath,amsfonts}
\usepackage{array}
\usepackage[caption=false,font=normalsize,labelfont=sf,textfont=sf]{subfig}
\usepackage{textcomp}
\usepackage{stfloats}
\usepackage{url}
\usepackage{verbatim}
\usepackage{graphicx}
\usepackage{cite}
\usepackage{amssymb}
\usepackage{algpseudocode}
\usepackage{tikz}
\usepackage{pgfplots}
\usepackage{xcolor}
\usepackage{booktabs}  
\usepackage{multirow}
\usepackage{hyperref}
\usepackage[ruled]{algorithm2e}
\SetAlCapFnt{\normalfont}
\SetAlCapNameFnt{\normalfont}
\SetKwSty{textbf}
\SetArgSty{textnormal}
\SetDataSty{textnormal}
\usepackage[T1]{fontenc}
\usepackage{newtxtext, newtxmath}

\pgfplotsset{compat=1.18}
\begin{document}

\title{EERLoss: A Novel Loss Function for Training Deep Biometric Models.
A Case Study in Keystroke Dynamics}

\author{Nahuel~Gonzalez\IEEEauthorrefmark{2}, 
        Marta~Robledo-Moreno\IEEEauthorrefmark{1}, 
        Ivan~DeAndres-Tame\IEEEauthorrefmark{1}, \\
        Ruben~Vera-Rodriguez\IEEEauthorrefmark{1}, 
        and~Ruben~Tolosana\IEEEauthorrefmark{1}
\thanks{\IEEEauthorrefmark{2}LSIA, University of Buenos Aires, Argentina (nahuel.gonzalez@lsia.fi.uba.ar)}
\thanks{\IEEEauthorrefmark{1}BiometricsAI, Universidad Autonoma de Madrid, Spain (\{marta.robledo, ivan.deandres, ruben.vera, ruben.tolosana\}@uam.es)}
\thanks{Manuscript received May 2026}}



\maketitle

\begin{abstract}
Deep learning approaches to biometric verification are commonly trained by optimizing indirect objectives, creating a misalignment between the optimization process and the primary evaluation metric, typically the Equal Error Rate (EER). This paper introduces \textit{EERLoss}: a subdifferentiable, arbitrarily accurate approximation to EER for training deep biometric models. Furthermore, this framework has the potential to be adapted to optimize any specific operating point on the DET curve, enhancing its generalizability. To validate this approach, \textit{EERLoss} is evaluated on a particularly demanding behavioral biometric modality: keystroke dynamics verification. This task is characterized by its high intra-class and low inter-class variability. Experiments are conducted on the large-scale KVC-onGoing benchmark, incorporating data from over 185,000 subjects across different scenarios. A comprehensive ablation study initially demonstrates the superiority of \textit{EERLoss} in comparison to existing state-of-the-art loss functions. It also converges substantially faster compared to other losses, reducing the overall training cost. Additionally, a comparison is made between the proposed loss and the KVC-winning architecture by re-training it with \textit{EERLoss}, demonstrating that the proposed approach significantly outperforms the original SoTA, achieving a relative EER reduction of up to approx. 30\%. This improvement on a challenging, large-scale benchmark validates the effectiveness of \textit{EERLoss} as a task-aligned training objective specifically suited for high-variance biometric traits.
\end{abstract}


\vspace{2mm}
\section{Introduction}
\label{sec:intro}

Biometric verification systems decide whether two samples belong to the same identity while trading off security and usability. The level of risk tolerance varies depending on the application \cite{jain2006biometrics}. For instance, high-security applications like banking logins prioritize the absolute exclusion of impostors, accepting occasional user lockouts as a necessary operational cost. Conversely, friction-sensitive scenarios, such as smartphone unlocking or continuous desktop re-authentication, prioritize a seamless user experience, tolerating transient false rejections that allow for immediate recovery. In this context, the Equal Error Rate (EER) denotes performance at the operating point where the False Acceptance Rate (FAR) is equal to the False Rejection Rate (FRR). A lower EER is indicative of a system that effectively admits genuine users and rejects impostors.

Deep biometric models are commonly trained to optimize for embedding separability, creating a fundamental misalignment with the common evaluation standard, typically the EER. Current systems rely on diverse indirect objectives, including cross-entropy and various margin-based metric losses such as contrastive \cite{chopra2005contrastive}, triplet \cite{schultz2003triplet}, and sophisticated angular losses like ArcFace, CosFace or AdaFace \cite{deng2019arcface, wang2018cosface, kim2022adaface}. While these methods successfully enhance the clustering of embeddings, their primary goal is to maximize the distance between classes. This inherent mismatch between the training objective (separability) and the evaluation metric (EER) can result in sub-optimal performance, particularly when the decision boundary is critical to the security and usability trade-off.

\vspace{2mm}
The present study aims to address this gap by introducing \textit{EERLoss}, an arbitrarily accurate, subdifferentiable approximation of the EER for training deep biometric models. Unlike conventional losses, \textit{EERLoss} aligns the training objective directly with the evaluation metric, optimizing the actual operational performance of a biometric system. The proposed loss function also has the potential to be adapted to optimize any specific operating point on the DET curve, enhancing its generalizability. To empirically validate \textit{EERLoss}, this work focuses on the challenging domain of behavioral biometrics, specifically keystroke dynamics (KD) \cite{shadman2025keystrokesurvey}. This modality is notorious for its high temporal variability, noise, and data sparsity. Demonstrating robust optimization of EER in this complex, large-scale scenario provides strong evidence of the method’s efficacy and its potential for generalization to other biometric traits, such as gait \cite{delgado_gait} or signature verification \cite{tolo_deep, maiorana2025signature}.

\vspace{2mm}
\noindent Our main contributions are:
\begin{itemize}
    \item[-] The formulation of a novel loss function named \textit{EERLoss}: a subdifferentiable, and arbitrarily accurate approximation of the EER metric, enabling direct end-to-end optimization.

    \item[-] The code's implementation of EERLoss has been made available\footnote{\href{https://github.com/BiDAlab/EERLoss}{https://github.com/BiDAlab/EERLoss}}. 

    \item[-] A comprehensive comparison study that validates the design choices of \textit{EERLoss} and compares its performance and training time against multiple SoTA loss functions.

    \item[-] A new state-of-the-art result on the large-scale KVC-onGoing benchmark, where \textit{EERLoss} significantly outperforms the previous best method, achieving a relative EER reduction of up to 30.7\%.
\end{itemize}

The remainder of the paper is structured as follows. Sec.~\ref{sec:relatedwork} provides an overview of loss functions for biometric models and keystroke dynamics for biometric verification. The problem statement is described in Sec.~\ref{sec:problemStatement}. Sec.~\ref{sec:definitions} presents the necessary definitions for the proposed loss function, which is further elaborated in Sec.~\ref{sec:loss}. The experimental setup is described in Sec.~\ref{sec:setup}, while the obtained results are presented in Sec.~\ref{sec:results}. In Sec.~\ref{sec:theo}, it is provided a theoretical justification for the performance gains of the proposed loss function. Finally, Sec.~\ref{sec:conclusoin} summarizes the findings and outlines directions for future work.

\section{Related Work}
\label{sec:relatedwork}

This section provides a review of the most relevant literature to contextualize the need for the proposed loss function. While the present study is focused on the use of KD for biometric verification, the proposed formulation is general and can be applied to other biometric tasks.

\subsection{Loss Functions for Biometric Models}
In the context of deep learning, the loss function defines the optimization target that guides the updating of model parameters during training. Meanwhile, performance metrics quantify the ability of the trained model to generalize to unseen data. The selection of an appropriate loss function is essential, as it determines the extent to which a model aligns its internal representations with the final evaluation criteria.

As stated in \cite{terven2025comprehensive}, two key properties of a well-defined loss function are differentiability (or subdifferentiability, as discussed in Sec.~\ref{sec:problemStatement}) and smoothness (meaning that is has a continuous gradient without abrupt discontinuities that could destabilize training).

In computer vision and biometric recognition, several well-established loss functions have proven effective for learning discriminative embeddings, including Contrastive loss~\cite{chopra2005contrastive}, Triplet loss~\cite{schultz2003triplet}, CosFace~\cite{wang2018cosface}, ArcFace~\cite{deng2019arcface},
AdaFace \cite{kim2022adaface} or Set2Set~\cite{gonzalez2024type2branch}. These formulations promote intra-class compactness and inter-class separation in the embedding space, indirectly improving verification metrics such as accuracy, precision, recall, or F1-score. However, none of these losses explicitly target the EER, which remains one of the most relevant metric in biometric verification systems.

\subsection{Keystroke Dynamics for Biometric Verification}
Keystroke dynamics refers to the analysis of a person's typing behavior used to infer or verify identity. As summarized in the recent comprehensive survey by Shadman \textit{et al.}~\cite{shadman2025keystrokesurvey}, KD has become one of the most promising behavioral biometric modalities due to its low acquisition cost, non-intrusive nature, and seamless integration with existing human-computer interfaces. In contrast to physiological biometrics, such as fingerprint or face recognition, which require dedicated sensors, KD can operate continuously in the background during normal interaction without the need for additional sensors.

The KVC-onGoing Challenge\footnote{\href{https://sites.google.com/view/bida-kvc/}{https://sites.google.com/view/bida-kvc/}} ~\cite{stragapede2025kvc} has become the reference benchmark for keystroke-based biometric verification. It provides a unified experimental framework with a standard evaluation protocol, large-scale datasets, and consistent performance metrics, allowing researchers to benchmark new systems under identical conditions. The challenge is based on the Aalto University Keystroke Databases~\cite{dhakal2018aaltodesktop}\cite{palin2019aaltomobile}, which are the largest public collections of typing data to date. These databases include over 185,000 subjects across desktop and mobile scenarios. These datasets capture transcript-level typing sessions in natural conditions, enabling realistic modeling of behavioral variability while maintaining demographic diversity and sufficient enrollment samples per user. In contrast to earlier KD datasets (e.g., GREYC~\cite{2009greyc}, RHU~\cite{2014rhu}, Clarkson II~\cite{2017clarkson}, HuMIdb~\cite{BeCAPTCHA_BotDetectionHuMiDB}), the Aalto database introduced massive scale and cross-device realism.

The KVC-onGoing challenge establishes the current state of the art. It reports the best global EER performances of 3.33\% for the desktop task and 3.61\% for the mobile task. An analysis of the competing methods reveals a critical trend: the vast majority of systems rely on standard margin-based metric losses, such as triplet loss. The systems that demonstrated the highest levels of performance, LSIA and VeriKVC, exhibited distinct characteristics that set them apart from their competitors. The winning team from the LSIA employed a dual-branch architecture, \textit{i.e.}, Convolutional Neural Network (CNN) + Recurrent Neural Network (RNN) with Attention, which was trained using a novel custom loss function called Set2Set \cite{gonzalez2024type2branch} that extends the SetMargin \cite{morales2022setmargin} loss. The VeriKVC team (runner-up) employed a pure CNN architecture that was trained with the ArcFace loss function. In \cite{stragapede2025kvc}, Stragapede \textit{et al.} concluded that this focus on loss was a key factor, indicating that the selection of an optimal training objective is crucial. 

In this context, the KVC-onGoing is as an appropriate and rigorous benchmark for our proposed loss function, as it is precisely in these high-variability scenarios where a loss function that directly optimizes for the EER can demonstrate a measurable advantage over conventional methods that only optimize for separability.

\section{Problem Statement}
\label{sec:problemStatement}

A distance metric learning model computes embeddings for the labeled samples in a training batch. For each training batch, we define $L$ as the set of distances between embeddings of samples that share the same label; that is, samples originated from the same user. The set $I$ will consist of the distances between the embeddings of samples with different labels; that is, samples from different users.

Informally, our objective is to \emph{build a loss function that outputs the EER of such training batch} using $L$ and $I$ as input. However, despite the apparent simplicity of this task, it is inherently constrained by the need for loss functions to be subdifferentiable. As stated before, the EER is defined as the specific point at which the FAR and FRR intersect. Unfortunately, the operation of finding the intersection of generic curves is inherently non-differentiable. Given this consideration, we must abandon the idea of calculating the EER exactly within a loss function, leading to the following.

\noindent \textbf{Problem statement.} \emph{Find a subdifferentiable function $\mathcal{L}(L, I)$ that, given the sets $L$ and $I$ containing the distances between samples of the same user and those of different users, respectively, approximates $EER(L,I)$ arbitrarily well.} \\

\section{Definitions}
\label{sec:definitions}
Let $d$ denote a positive real number. Concretely, $d$ (sometimes with subscripts) will have two possible denotations: (i) the distance between the embedding vectors of a pair of samples, as calculated by a neural network, and (ii) the uniform global threshold of a distance metric learning (DML) verification system. We can assume that $d$ is bounded; that is, there is a constant $M$ such that $0 \leq d \leq M$. 

We now define the legitimate and impostor distance sets: 
\begin{align}
L = \{d^L_1 , \ldots , d^L_n \} \\
I = \{d^I_1 , \ldots , d^I_m \} 
\end{align}
where $n = |L|$, $m = |I|$, and $d^L_i, d^L_j \in \mathbb{R}^+$. In concreteness, the sets $L$ and $I$ will correspond to the distances between the embedding vectors of the samples of the same user and of different users in a training batch, respectively. 

From $L$, $I$, and a given threshold $d$, we can calculate the False Rejection Rate (FRR) and False Acceptance Rate (FAR) for a batch, in the form:
\begin{align}
FRR(L,d) &= 1 - \frac{|\{d^L \in L \, : \, d^L > d \}|}{|L|} \label{eq:FRR} \\
FAR(I,d) &= \frac{|\{d^I \in I \, : \, d^I < d  \}|}{|I|} \label{eq:FAR}
\end{align}
Note that $FRR(L,d)$ is a step function, monotonously decreasing with $d$. $FAR(I,d)$ is also a step function, but monotonously increasing. Thus, there is a unique value $d_{EER}$ in $[0, M]$ such that
\begin{equation}
FRR(L,d_{EER}) \, \leq \, FAR(I,d_{EER})
\end{equation}
and
\begin{equation}
FRR(L,d) \, > \, FAR(I,d) \,\,\,\,\,\,\,\, \forall d \, < \, d_{EER}
\end{equation}
Using the above, we define the \emph{batch EER} as
the average of the FAR and FRR at $d_{EER}$.

In the algorithms that follow, $K$ is a constant that will control the quality of the approximation. The error bound will decrease as $K$ increases, approaching zero as $K$ goes to infinity.  In practical cases we can have $K=1000$. The subscripts $L$ and $R$ (as in $M_L$, $M_R$, or $I_L$, $I_R$, or $d_L, d_R$) will denote the \emph{left} and \emph{right} of $d_{EER}$, respectively.

\section{EERLoss: Proposed Loss Function}
\label{sec:loss}

We will find $\mathcal{L}(L,I)$ as in the Problem Statement by building the objective function from the bottom up. We will first show, in Section~\ref{subsec:smoothFARFRR}, how to approximate the FAR and FRR of a batch at a given threshold in a subdifferentiable way. Using these approximations, in Section~\ref{subsec:smoothBS} we will estimate $d_{EER}$ through a subdifferentiable binary search over a reasonably chosen interval. Once $d_{EER}$ is determined, the calculation of the EER of the batch to be used as an objective function is immediate.

However, the EER is a local, stepwise objective that fails to encourage the network to separate samples when their distance is far from $d_{EER}$. To overcome this problem, we will propose an alternative formulation that offers a smooth target extending to the entire area of overlap between the FAR and FRR curves. In addition, we will explore adding a margin away from $d_{EER}$ and delinearizing the contribution of distant values.

\subsection{Smooth FAR and FRR}
\label{subsec:smoothFARFRR}

Given a positive $d$, together with sets $L$ and $I$ as in Section~\ref{sec:definitions}, the FAR and FRR of the batch at threshold $d$ can be approximated using Algorithms~\ref{alg:FRR} and~\ref{alg:FAR}, which are subdifferentiable renditions of equations (\ref{eq:FRR}) and (\ref{eq:FAR}). Our points of departure are the following smoothings of the equations for FAR and FRR. Let
\begin{align}
M_R(d, d^L_i) &= \tanh{[K.(\max{\{d, d^L_i\}} - d)]} \\
M_L(d, d^I_j) &= \tanh{[K.(d - \min{\{d, d^I_j\}})]}
\end{align}
and define the smooth counterparts of~\ref{eq:FRR} and~\ref{eq:FAR} as
\begin{align} 
FRR_S(L,d) &= 1 - \frac{1}{|L|}\sum_{d^L_i \in L} M_R(d, d^L_i) \label{eq:smoothFRR} \\
FAR_S(I,d) &= \frac{1}{|I|} \sum_{d^I_j \in I} M_L(d, d^I_j) \label{eq:smoothFAR}
\end{align}
To understand how equation (\ref{eq:smoothFRR}) provides a smooth approximation to equation (\ref{eq:FRR}), observe that 
\begin{align}
\max{\{d, d^L_i\}} - d = 
\begin{cases}
0 & \text{if } d^L_i \leq d \\
> 0 & \text{if } d^L_i > d
\end{cases}
\end{align}
Hence, for large $K$, we have that
\begin{align}
M_R(d, d^L_i) = 
\begin{cases}
\approx 0 & \text{if } d^L_i \leq d \\
\approx 1 & \text{if } d^L_i > d
\end{cases}
\end{align}
Thus,
\begin{equation}
FRR_S(L,d) \approx FRR(L,d)
\end{equation}
and identical reasoning yields
\begin{equation}
FAR_S(I,d) \approx FAR(L,d)
\end{equation}
The FRR algorithm~\ref{alg:FRR} is a direct tensorial implementation of equation (\ref{eq:smoothFRR}); the only difference is that it multiplies the result by 100 to scale it to a human-readable percentage. The FAR algorithm~\ref{alg:FAR} implements equation (\ref{eq:smoothFAR}) similarly.

\begin{algorithm}
\caption{Smooth False Rejection Rate ($FRR_S$)}
\label{alg:FRR}
\KwData{$L, d \geq 0$}
\KwResult{$frr$}
\vspace{1mm}
\hrule
\vspace{1mm}
$L_R \hspace{3mm} \gets max(d, L) - d$ 

$M_R \hspace{2mm} \gets tanh(K * L_R)$ 

\textbf{return} $100 * avg(M_R)$ \\
\end{algorithm}
\begin{algorithm}
\caption{Smooth False Acceptance Rate ($FAR_S$)}
\label{alg:FAR}
\KwData{$I, d \geq 0$}
\KwResult{$far$}
\vspace{1mm}
\hrule
\vspace{1mm}
$I_L \hspace{4mm} \gets d - min(d, I)$ 

$M_L \hspace{2mm} \gets tanh(K * I_L)$ 

\textbf{return} $100 * avg(M_L)$ \\
\end{algorithm}

Note that using the $tanh$ function both in equations (\ref{eq:smoothFRR}) and (\ref{eq:smoothFAR}), as well as in Algorithms~\ref{alg:FRR} and~\ref{alg:FAR}, is not mandatory. Any continuous, monotonously increasing function $f$ such that $f(0) = 0$ and $f(x) \to 1$ fast enough when $x \to \infty$ should suffice for the purpose. In particular, $tanh$ was chosen for behaving well when training neural networks, and having fast tensorial implementations in modern GPUs

The proof that Algorithms~\ref{alg:FRR} and~\ref{alg:FAR} are subdifferentiable is almost trivial. As subtraction, multiplication, $min$, $tanh$, and $avg$ are  subdifferentiable, and compositions of subdifferentiable functions are subdifferentiable, we have that both algorithms compute subdifferentiable approximations to the FAR and FRR of the batch at a threshold $d$. The proofs are similar for the algorithms that follow. Therefore, we will not repeat them.

\subsection{A Smooth Implementation of Binary Search}
\label{subsec:smoothBS}

We now want to use the proposed FAR and FRR approximations to perform a binary search for $d_{EER}$. The subdifferentiable binary search step is shown in Alg.~\ref{alg:BSSDERR}. It takes for input an interval $(d_L, d_R)$, where $d_L < d_{EER} < d_R$, and returns a similar interval $(d'_L,d'_R)$ of half the size. 

\begin{algorithm}
\caption{Binary Search Step for $d_{EER}$ (BSS)}
\label{alg:BSSDERR}
\KwData{$L, I, d_L < d_R$}
\KwResult{$d'_L < d'_R$}
\vspace{1mm}
\hrule
\vspace{1mm}
$d \hspace{2.7mm} \gets (d_L + d_R) / 2$ 

$far \hspace{0.7mm} \gets FAR_S(I, d)$ 

$frr \hspace{1mm} \gets FRR_S(L, d)$ 

$m \hspace{2mm} \gets max\{far, frr \}$ 

$c_L \hspace{2.5mm} \gets exp(K * (frr - m))$ 

$c_R \hspace{2.5mm} \gets exp(K * (far - m))$ 

$d'_L \hspace{1.5mm} \gets d_L * (1 - c_L) + d * c_L$

$d'_R \hspace{1.5mm} \gets d_R * (1 - c_R) + d * c_R$ 

\textbf{return} $d'_L, d'_R$
\end{algorithm}

To understand the correctness of the algorithm, note that $c_L$, calculated in line 5, acts as a flag signaling whether the midpoint $d$ is to the left of $d_{EER}$ or not. If $d \leq d_{EER}$, then $frr \geq far$. Hence
$$frr - max(far,frr) = 0$$
yielding
$$c_L = exp(K * 0) = 1$$
whereas if $d > d_{EER}$, then $frr < far$ and
$$frr - max(far,frr) < 0$$
Thus, $c_L \approx 0$ because the term inside the exponential is negative and large. Similarly, $c_R$ marks whether $d \geq d_{EER}$ is true or not. 

Using this subdifferentiable binary search step, Alg.~\ref{alg:BS} implements
a subdifferentiable binary search for $d_{EER}$ given sets $L$ and $I$. The third parameter $n$ is a positive integer that controls the number of steps. 

\begin{algorithm}
\caption{Binary Search for $d_{EER}$ (DEER)}
\label{alg:BS}
\KwData{$L, I, n > 0$}
\KwResult{$d_{EER}$}
\vspace{1mm}
\hrule
\vspace{1mm}
$d_L \gets 0.5 * avg(L)$ 

$d_R \gets 1.5 * avg(I)$ 

\For{$i \leq n$}{
    $d_L, d_R \gets BSS(d_L, d_R)$ \\
}
$\textbf{return} \,\, (d_L + d_R) / 2$ 
\end{algorithm}

Note that the initial endpoints are based on the average of the distance values in $L$ and $I$. The objective is to make the endpoints less sensitive to a choice of constants or to any variations in the embeddings. During the first training epoch, it is possible that the average distance between embeddings of samples of the same user is larger than the average distance between embeddings of samples of different users. Thus, the 0.5 and 1.5 constants are used to make sure that $d_L < d_R$ in the initial search interval.

\subsection{Encoding the EER as a Loss Function}

We have all the building blocks needed to encode the EER as a loss function, which is shown in Alg.~\ref{alg:EERLOSS}. We simply estimate $d_{EER}$ with Alg.~\ref{alg:BS}, compute the FRR and FAR at $d_{EER}$ using Algorithms~\ref{alg:FRR} and~\ref{alg:FAR}, take the EER to be its average, and use it as the loss value. This algorithm fulfills the Problem Statement of Section~\ref{sec:problemStatement}.
\begin{algorithm}
\caption{EER Loss}
\label{alg:EERLOSS}
\KwData{$L, I, n > 0$}
\KwResult{$eer$}
\vspace{1mm}
\hrule
\vspace{1mm}
$d_{EER} \gets DEER(L, I, n)$ 

\textbf{return} $[FAR_S(I,d_{EER}) + FRR_S(L,d_{EER})] / 2$
\end{algorithm}

Although the above algorithm encodes the EER objective directly, it is not optimal for training a neural network. To see why, note that the EER value can only be reduced by eliminating false negatives and positives. This is an inherently discrete process that doesn't provide a clear target for optimization when no point can be made to cross the $d_{EER}$ threshold.

\subsection{Area-Based Formulation of the Loss Function}

To overcome the limitation discussed in the previous subsection, we propose a loss function that measures the area of overlap between the FAR and FRR curves. As described in Section~\ref{sec:definitions}, the FAR and FRR curves are stepwise, with steps of $1/|I|$ and $1/|L|$ respectively. Thus, letting $A_R(L)$ be the area of the FRR curve to the right of $d_{EER}$ and $A_L(I)$ the area of the FAR curve to the left of $d_{EER}$, we have
\begin{align}
A_R(L) &= \frac{1}{|L|} \sum_{\substack{d^L_i \in L \\ d^L_i > d_{EER}}} {(d^L_i - d_{EER})} \label{eq:eqARL} \\
A_L(I) &= \frac{1}{|I|}\sum_{\substack{d^I_j \in I \\ d^I_j < d_{EER}}} {(d_{EER} - d^I_j)} \label{eq:eqALI} 
\end{align}
Hence, the total area of overlap between the FAR and FRR curves is given by
\begin{equation}
\label{eq:ALI}
A(L,I) = A_R(L) + A_L(I)
\end{equation}
By itself, the area of overlap does not provide a good target for minimization, as it can be easily exploited by the neural network under training without improving the intended authentication performance. Preliminary experiments showed that, to avoid this undesirable behavior, the area of overlap needs to be scaled down by $d_{EER}$, giving
\begin{equation}
\label{eq:Larea}
\mathcal{L}_{AREA}(L,I) = \frac{A(L,I)}{d_{EER}}
\end{equation}
Alg.~\ref{alg:AREALOSS} shows a subdifferentiable tensorial implementation of the above equation. Unlike the previous algorithms, this one is straightforward and requires no further explanation.

\begin{algorithm}
\caption{Area-based Loss}
\label{alg:AREALOSS}
\KwData{$L, I, n > 0$}
\KwResult{$area$}
\vspace{1mm}
\hrule
\vspace{1mm}
$d_{EER} \gets DEER(L,I,n)$ 

$L_R \gets max(0, L - d_{EER})$ 

$A_R \gets sum(L_R) \, / \, |L| $ 

$I_L \gets max(0, d_{EER} - I)$ 

$A_L \gets sum(I_L) \, / \, |I| $ 

\textbf{return} $(A_R + A_L) / d_{EER}$
\end{algorithm}
\vspace{-6mm}
\subsection{Generalizations}

It is customary for distance metric learning loss functions to include a parameter $\alpha$ that enforces a larger separation between positive and negative samples. For example, the well-known TripletLoss \cite{chechik2010large} and, for keystroke dynamics verification, the recent SetMarginLoss \cite{morales2022setmargin} and Set2Set loss \cite{gonzalez2024type2branch}. Extending equations (\ref{eq:eqARL}) and (\ref{eq:eqALI}) to include a margin is straightforward.

Note that in the aforementioned equations, those values of $d^L_i$ and $d^I_j$ further away from the thresholds contribute more to the sums, linearly. This is not optimal; the efforts of the loss function are better spent focusing on values near the threshold. For this purpose, we introduce a constant $\beta$ such that $0 < \beta < 2$, to delinearize the contribution of points further from $d_{ERR}$, in the form
\begin{align}
A_R(L) &= \frac{1}{|L|} \sum_{\substack{d^L_i \in L \\ d^L_i > t_L(\alpha)}} \left(\frac{\max{\{\epsilon, d^L_i - d_{EER}\}}}{d_{EER}}\right)^\beta \\
A_L(I) &= \frac{1}{|I|} \sum_{\substack{d^I_j \in I \\ d^I_j < t_R(\alpha)}} \left(\frac{\max{\{\epsilon, d_{EER} - d^I_j}\}}{d_{EER}}\right)^\beta 
\end{align}
where the small $\epsilon$ term has been introduced to avoid arithmetic errors if the differences are too small. With the above modification, the $\alpha/\beta$ area-based loss function takes the form
\begin{equation}
\mathcal{L}_{AREA}(L,I)  = A_R(L)^{1/\beta} + A_L(I)^{1/\beta}
\end{equation}
Algorithm~\ref{alg:ALPHABETA} shows the subdifferentiable, tensorial implementation of the above formula. 

\begin{algorithm}
\caption{EERLoss ($\alpha, \beta$)}
\label{alg:ALPHABETA}
\KwData{$L, I, n > 0, \alpha > 0, \beta > 0$}
\KwResult{$\alpha/\beta$ area}
\vspace{1mm}
\hrule
\vspace{1mm}
$d_{EER} \gets DEER(L,I,n)$ 

$t_R \gets (1 - \alpha) * d_{EER}$

$L_R \gets pow(max(\epsilon, L - d_R) / d_{EER}, \,\beta)$ 

$A_R \gets pow(sum(L_R) \, / \, |L|, \, 1/\beta) $ 

$t_L \gets (1 + \alpha) * d_{EER}$

$I_L \gets pow(max(\epsilon, t_L - I) / d_{EER}, \,\beta)$ 

$A_L \gets pow(sum(I_L) \, / \, |I|, \, 1/\beta) $

\textbf{return} $A_L + A_R$
\end{algorithm}

\section{Experimental Setup}
\label{sec:setup}

In order to validate the proposed loss function, experiments are designed to isolate its impact and compare it directly against state-of-the-art methods. All experiments are conducted based on the KVC-onGoing benchmark \cite{stragapede2025kvc}. This benchmark uses the Aalto Desktop \cite{dhakal2018aaltodesktop} and Aalto Mobile \cite{palin2019aaltomobile} datasets, which together comprise data from over 185,000 subjects.

\subsection{Evaluation Protocol}
Our empirical evaluation protocol proceeds in two distinct stages: \textbf{1. Ablation Study (Sec.~\ref{sec:results_ablation}):}
The goal of this first phase is to efficiently select the optimal hyperparameters for the proposed loss described in Algorithm~\ref{alg:ALPHABETA}, \textit{i.e.}, \textit{EERLoss} ($\alpha, \beta$) and to conduct a fair comparison against other losses, in terms of performance and training time. This study is performed using a reduced-parameter version of the main architecture (as specified in Sec.~\ref{sec:model_arch}), trained on a subset of 1,000 users randomly sampled from the KVC-Desktop development set. Evaluation is then conducted on a fixed 1,000-user subset of the official KVC-Desktop evaluation set. \textbf{2. SoTA Comparison (Sec.~\ref{sec:results_sota}):}
The second stage provides a definitive, large-scale comparison of the winning configuration against the strongest published SoTA baseline, the \textit{Set2Set} loss \cite{gonzalez2024type2branch}. For this comparison, the full-capacity model architecture (Sec.~\ref{sec:model_arch}) is used. The model is trained on the entire KVC development sets for both Desktop (168k users) and Mobile (37k users) tasks. Evaluation is performed on the full, official KVC evaluation sets, strictly adhering to the official benchmark protocol.
    
The evaluation relies on the primary metrics defined by the KVC benchmark. The main ranking metric is the Global EER, which evaluates performance using a single, fixed decision threshold for all subjects. Additionally, the Mean per-subject EER (\textit{Avg. per-user EER}) is reported. This metric is critical as it calculates an optimal, user-specific threshold, reflecting a more realistic deployment scenario on personal devices. To analyze performance under varying enrollment conditions, both EER metrics are evaluated as a function of $G$, the number of enrollment samples, where $G \in \{1, 2, 5, 7, 10\}$. 

\subsection{Model Architecture}
\label{sec:model_arch}
To isolate the impact of the loss function, our experiments are based on two scaled versions of a single base architecture. We adopt the model from the LSIA team \cite{gonzalez2024type2branch}, which won the KVC-onGoing Challenge \cite{stragapede2025kvc}. This model is a dual-branch (RNN + CNN) embedding network. As detailed in \cite{gonzalez2024type2branch}, the recurrent branch consists of two bidirectional GRU layers with Self-Attention. The convolutional branch uses a stack of three 1D convolutions with progressively increasing filter counts ($F$, $2F$, $4F$), interspersed with Channel Attention modules. Both branches use Temporal Attention at their input. The outputs of both branches are concatenated and passed through a final MLP to produce a fixed 256-dimensional embedding. The capacity of this architecture is controlled by two hyperparameters: the width of the GRU and dense layers ($W$) and the base number of filters in the convolutional branch ($F$). We define the two configurations for our experiments: the reduced-capacity one ($W = 256$, $F = 128$) and the full-capacity one ($W = 512$, $F = 1024$).

\subsection{Baseline Loss Functions}
\label{sec:baselines}

We compare our proposed \textit{EERLoss} (both the direct formulation described in Alg.~\ref{alg:EERLOSS} and the refined version of Alg.~\ref{alg:ALPHABETA}) against a comprehensive set of SoTA and widely-used loss functions identified in the KVC analysis:

\begin{itemize}
    \item \textbf{Semi-Hard Triplet Loss \cite{schultz2003triplet}:} A foundational metric learning objective that minimizes the distance between an anchor and a positive sample while maximizing the distance to a negative sample. This was the most common baseline adopted by multiple teams in the KVC \cite{stragapede2025kvc}.
    \item \textbf{ArcFace \cite{deng2019arcface}:} A widely-adopted loss that introduces an \textit{additive angular margin} to the target logit, enforcing a more discriminative embedding space on the hypersphere. It was used by the VeriKVC team \cite{stragapede2025kvc}.
    \item \textbf{CosFace \cite{wang2018cosface}:} A related approach that introduces an \textit{additive cosine margin} to the target logits to improve intra-class compactness and inter-class separation.
    \item \textbf{Set2Set:} This is the custom loss function used by the KVC winning LSIA team \cite{gonzalez2024type2branch}. It is an extension of SetMargin Loss described in \cite{morales2022setmargin}.
\end{itemize}

\section{Results}
\label{sec:results}
\begin{table*}[t!]
\caption{Comparative study results for different loss functions and their hyperparameters. Parameter settings are described in the text, and P denotes the number of distinct users (sets) per batch. Alg.~\ref{alg:EERLOSS} refers to the direct formulation of the proposed loss, while Alg.~\ref{alg:ALPHABETA} is the refined formulation. We compare the Global EER (\%) and the Average per-user EER (\%) varying the number of enrollment samples (G).}

\setlength{\lightrulewidth}{0.2pt}
\centering
\resizebox{0.99\textwidth}{!}{
\begin{tabular}{ll c ccccc ccccc}
\toprule
\multicolumn{1}{c}{\multirow{2}{*}{\textbf{Loss}}} & \multicolumn{1}{c}{\multirow{2}{*}{\textbf{Parameters}}} & \multicolumn{1}{c}{\multirow{2}{*}{\textbf{P}}} & \multicolumn{5}{c}{\textbf{Global EER (\%)}} & \multicolumn{5}{c}{\textbf{Avg. per-user EER  (\%)}} \\
\cmidrule(lr){4-8} \cmidrule(lr){9-13}
 & & & G=1 & G=2 & G=5 & G=7 & G=10 & G=1 & G=2 & G=5 & G=7 & G=10 \\
\toprule
Semi-Hard Triplet \cite{schultz2003triplet} & - & - & 8.28 & 6.75 & 5.67 & 5.60 & 5.35 & 6.92 & 5.83 & 5.27 & 5.29 & 6.04 \\
\midrule
\multirow{2}{*}{ArcFace \cite{deng2019arcface}} & $m=0.2; s=16$ & 50 & 13.62 & 12.05 & 10.89 & 10.38 & 10.20 & 10.50 & 9.03 & 7.97 & 7.10 & 7.03 \\
 & $m=0.1; s=8$ & 50 & 14.30 & 12.75 & 11.59 & 11.35 & 10.84 & 10.62 & 9.35 & 8.41 & 7.45 & 7.73 \\
\midrule
CosFace \cite{wang2018cosface} & $m=0.1; s=8$ & 50 & 13.05 & 11.67 & 10.23 & 9.97 & 9.70 & 10.28 & 8.73 & 8.02 & 6.85 & 7.53 \\
\midrule

Set2Set \cite{gonzalez2024type2branch} & - & - & 8.02 & 6.60 & 5.47 & \textbf{5.26} & \textbf{4.70} & 6.92 & 5.73 & 4.86 & 4.75 & 5.03 \\
\midrule

EERLoss (Alg.~\ref{alg:EERLOSS}) & - & 40 & 13.63 & 11.98 & 10.71 & 10.20 & 9.99 & 12.19 & 10.67 & 9.37 & 9.07 & 9.73 \\
\midrule
\multirow{4}{*}{EERLoss (Alg.~\ref{alg:ALPHABETA})} & $\beta=1$ & 40 & 8.20 & 6.81 & 5.84 & 5.70 & 5.58 & 6.83 & 5.67 & 4.94 & 4.91 & 5.37 \\
 & $\beta=0.85$ & 10 & 11.83 & 10.10 & 8.77 & 8.49 & 8.42 & 10.04 & 8.22 & 6.84 & 5.97 & 5.94 \\
 & $\beta=0.85$ & 40 & 8.58 & 6.86 & 6.00 & 5.71 & 5.54 & 6.89 & 5.39 & 4.68 & \textbf{3.86} & \textbf{3.64} \\
 & $\beta=0.85$; data aug. & 40 & \textbf{7.74} & \textbf{6.27} & \textbf{5.41} & 5.29 & 4.80 & \textbf{6.51} & \textbf{5.07} & \textbf{4.51} & 4.28 & 4.61 \\
\bottomrule
\end{tabular}%
}
\label{tab:main_ablation}
\end{table*}

In this section, we present the empirical results of our proposed method. We first conduct an comparative study to determine the optimal configuration of \textit{EERLoss} (Sec.~\ref{sec:results_ablation}). We then compare our optimized configuration against the SoTA baseline on the full KVC benchmark (Sec.~\ref{sec:results_sota}).

\subsection{Comparative Study and Hyperparameter Selection}
\label{sec:results_ablation}
 
\begin{figure}[ht]
    \centering
    \scalebox{0.9}{
        \pgfplotsset{compat=newest}




  \begin{tikzpicture}
    \begin{axis}[
        xmajorgrids,
        ymajorgrids,
        zmajorgrids,
        xmin=0.5,
        xmax=1.5,
        ymin=0.0,
        ymax=0.05,
        zmin=2,
        view={-25}{30},
        mesh/ordering=y varies,
        mesh/cols=3,
        mesh/rows=6,
        mesh/check=false,
        xlabel={$\beta$},
        xtick={0.5,0.6,0.7,0.8,0.9,1.0,1.1,1.2,1.3,1.4,1.5},
        ylabel={$\alpha$},
        zlabel={EER [\%]},
        xticklabel style={rotate=-40},
        ytick={0.0,0.01,0.02,0.03,0.04,0.05},
        yticklabels={0.0,0.01,0.02,0.03,0.04,0.05},
        yticklabel style={rotate=40,
                /pgf/number format/fixed, 
                /pgf/number format/precision=2,
                ytick scale label code/.code={},
                xshift=-0.5cm,
                yshift=0.1cm
            },        
        colormap={greenred}{
            color(0cm)=(green);
            color(1cm)=(yellow);
            color(6cm)=(red)
        },
        point meta min=2.0, 
        point meta max=4.5, 
        colorbar style={
            ytick={2.0,2.5,3.0,3.5,4.0}
        }        
        ]
      \addplot3[surf, 
         shader=flat]
      table {3DPLOT.csv};
      
    \end{axis}
    
  \end{tikzpicture}

}
    \caption{Average per-user EER (interpolated) for different values of $\alpha$ and $\beta$ in Alg.~\ref{alg:ALPHABETA}. The minimum is achieved near $\alpha=0, \beta=0.85$.}
    \label{fig:alphabeta-3D}
\end{figure}
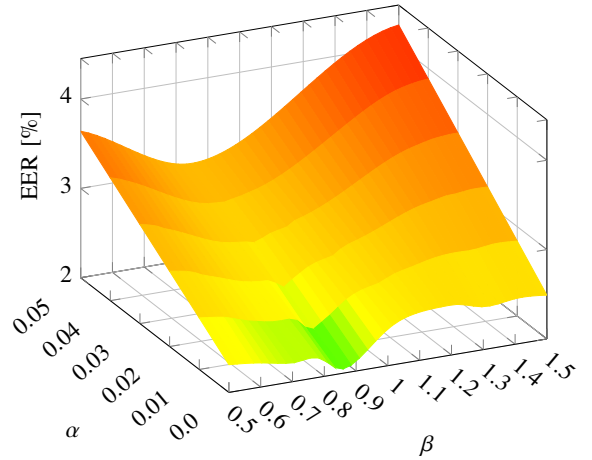

Our \textit{EERLoss} variant (Alg.~\ref{alg:ALPHABETA}) is controlled by the parameters $\alpha$ and $\beta$. We first perform a grid search to find the optimal combination of these parameters. Figure~\ref{fig:alphabeta-3D} visualizes the resulting average per-user EER surface. We observe a clear global minimum achieved near $\alpha=0$ and $\beta=0.85$.

Table~\ref{tab:main_ablation} details the configurations for all evaluated loss functions. The \textit{Parameters} column denotes the optimal hyperparameters employed (even though more configurations where investigated), such as the margin ($m$) and scale ($s$) for the ArcFace and CosFace baselines, and the $\beta$ parameter for our \textit{EERLoss}. The $P$ column indicates the number of distinct users (sets) per batch, a value maximized to fit GPU capacity, as higher $P$ is better. The preliminary findings indicate that standard losses are highly sensitive and frequently suboptimal in this task. However, Semi-Hard Triplet loss has been demonstrated to provide a stable baseline performance.

Our proposed \textit{EERLoss} demonstrates a significantly superior performance. While the base direct EER approximation of Alg.~\ref{alg:EERLOSS} is not competitive, the \textit{EERLoss} ($\beta=0.85, K=40$) variant of Alg.~\ref{alg:ALPHABETA} outperforms all baselines. This enhancement is refined through the implementation of a data augmentation technique, where Gaussian noise with a 10 millisecond variance is introduced to the keystroke timing features, simulating minor temporal jitter. With this final configuration, we achieve the best overall Global EER: 7.74\% in the hardest enrollment scenario ($G=1$).

However, we argue that the \textit{Avg. per-user EER} is a more significant and practical metric for this application. Biometric verification on personal devices (desktop or mobile) naturally supports user-specific decision thresholds rather than a single global one. In this more realistic evaluation setting, the proposed system (without data augmentation) achieves an average per-user EER of 3.64\% for $G=10$ enrollment samples and 3.86\% for $G=7$. This is a remarkable result, especially given that this comparative study was performed on a reduced dataset with a smaller-capacity model.

\begin{figure}[t]
    \centering
    \includegraphics[width=0.8\linewidth]{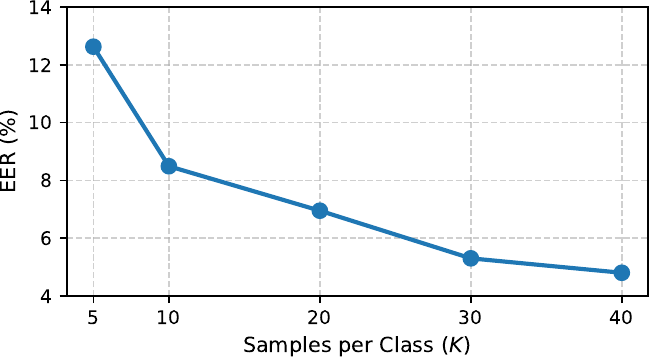}
    \vspace{-2mm}
    \caption{Effect of Batch Size. Error decreases consistently as the number of samples per class grows, yet the method remains stable even at 5 samples per batch.}
    \label{fig:impactbatch}
\end{figure}

To further validate the stability of the proposed objective, we analyzed the impact of batch composition, in particular, the number of samples per class. As illustrated in Figure \ref{fig:impactbatch}, while the error rate consistently decreases as the number of samples per class increases (due to a better approximation of the global distribution), EERLoss converges effectively and remains stable even with minimal sampling (e.g., 5). This robustness to batch sparsity represents a significant operational advantage over traditional margin-based losses, which often require large, carefully mined batches to approximate global class centers.

\subsection{Comparison with State-of-the-Art}
\label{sec:results_sota}

Following the comparative study, we compare our optimized model using the proposed loss \textit{EERLoss} with $\alpha=0, \beta=0.85, P=40$, against the same model with the strongest SoTA baseline, the \textit{Set2Set} loss. Both methods are trained from scratch on the full KVC-onGoing development sets (Desktop and Mobile) using the full-capacity model architecture, as described in Sec.~\ref{sec:model_arch}, to ensure a fair and direct comparison.

The final results are presented in Table~\ref{tab:sota_results}. Our proposed \textit{EERLoss} outperforms the \textit{Set2Set} baseline across both challenging, large-scale scenarios. On the \textbf{Desktop Task}, the performance gains are substantial. The proposed method achieves a Global EER of \textbf{1.15\%} (at $G=10$), which corresponds to a 30.7\% relative reduction in error compared to the 1.66\% achieved by the Set2Set baseline. The improvement in the more practical Avg. per-user EER is even more pronounced. We achieve \textbf{0.92\%} (at $G=10$) and \textbf{0.87\%} (at $G=7$), demonstrating a clear advantage over the baseline's 1.29\% and 1.27\% (both with a relative reduction of approx. 30\%). On the \textbf{Mobile Task}, the efficacy of our proposed method is also evident, outperforming the baseline in four out of the five enrollment scenarios for the Avg. per-user EER. It is notable that \textit{EERLoss} shows a particularly strong performance advantage in the few-shot enrollment scenarios ($G=1$ and $G=2$). For example, in the \textbf{Desktop task} at $G=1$, we achieve a Global EER of \textbf{2.74\%} vs. 3.20\% for \textit{Set2Set}. This suggests that the proposed method produces a more robust and well-structured embedding space, which allows for a more accurate definition of a user's identity from fewer samples.

\begin{table*}[t]
\centering
\caption{Quantitative comparison of our proposed \textit{EERLoss} against the \textit{Set2Set} (SoTA) baseline on the full KVC Mobile and Desktop tasks. We report Global EER (\%) and Avg. per-user EER (\%) as a function of the number of enrollment samples (G).}
\small
\begin{tabular}{ll ccccc ccccc}
\toprule
\multicolumn{1}{c}{\multirow{2}{*}{\textbf{Scenario}}} & \multicolumn{1}{c}{\multirow{2}{*}{\textbf{Loss}}} & \multicolumn{5}{c}{\textbf{Global EER (\%)}} & \multicolumn{5}{c}{\textbf{Avg. per-user EER (\%)}} \\
\cmidrule(lr){3-7} \cmidrule(lr){8-12}
 & & G=1 & G=2 & G=5 & G=7 & G=10 & G=1 & G=2 & G=5 & G=7 & G=10 \\
\midrule
\multirow{2}{*}{\textbf{Mobile}} & Set2Set \cite{gonzalez2024type2branch} & 3.44 & \textbf{2.53} & 2.11 & \textbf{2.02} & \textbf{2.03} & 2.41 & 1.97 & 1.49 & \textbf{1.51} & 1.71 \\
 & EERLoss  (Alg.~\ref{alg:ALPHABETA}) & \textbf{3.35} & \textbf{2.53} & \textbf{2.04} & 2.03 & 2.04 & \textbf{2.31} & \textbf{1.68} & \textbf{1.42} & 1.57 & \textbf{1.60} \\
\addlinespace
\multirow{2}{*}{\textbf{Desktop}} & Set2Set \cite{gonzalez2024type2branch}
 & 3.20 & 2.36 & 1.80 & 1.69 & 1.66 & 2.36 & 1.65 & 1.34 & 1.27 & 1.29 \\
 & EERLoss (Alg.~\ref{alg:ALPHABETA}) & \textbf{2.74} & \textbf{1.79} & \textbf{1.37} & \textbf{1.29} & \textbf{1.15} & \textbf{2.05} & \textbf{1.43} & \textbf{0.92} & \textbf{0.87} & \textbf{0.92} \\
\bottomrule
\end{tabular}%
\label{tab:sota_results}
\end{table*}

\subsection{Computational Efficiency and Parallelism}
\label{sec:efficiency}

\begin{table}[ht]
\caption{Training time comparison. Exp. 1 (Comparison Study) was run on an NVIDIA RTX 3070. Exp. 2 (SOTA Comparison - Desktop Task) was run on an NVIDIA A100.}
\centering
\small
\begin{tabular}{llr}
\toprule
\textbf{Experiment} & \textbf{Loss Function} & \textbf{Training Time} \\
\midrule
\multirow{5}{*}{Exp. 1 (\ref{sec:results_ablation})} & Set2Set \cite{gonzalez2024type2branch} & 4h 46min \\
 & Semi-Hard Triplet \cite{schultz2003triplet} & 2h 7min \\
 & CosFace \cite{wang2018cosface} & 46 min \\
 & ArcFace \cite{deng2019arcface} & 42 min \\
 & \textbf{EERLoss (Alg.~\ref{alg:ALPHABETA})} & \textbf{23 min} \\
\midrule
\multirow{2}{*}{Exp. 2 (\ref{sec:results_sota})} & Set2Set \cite{gonzalez2024type2branch} & 92h 47min \\
 & \textbf{EERLoss (Alg.~\ref{alg:ALPHABETA})} & \textbf{6h 52min} \\
\bottomrule
\end{tabular}

\label{tab:training_times_combined}
\end{table}

Beyond verification accuracy, computational cost is a critical factor for training large-scale models. Table~\ref{tab:training_times_combined} reports the total training time required for our two experimental stages. The comparison study (Exp. 1) was conducted on an NVIDIA RTX 3070. Due to the significant computational demands of training on the full-scale datasets, the SoTA comparison (Exp. 2) was performed on a NVIDIA A100. All experiments used the AdamW optimizer with an initial learning rate of $1 \times 10^{-4}$ and a patience of 40 epochs. In both settings, the proposed \textit{EERLoss} is dramatically more efficient. In Exp. 1, \textit{EERLoss} converges in only 23 minutes (vs. 4h 46min for \textit{Set2Set}). This efficiency scales directly to the large-scale task (Exp. 2), where \textit{EERLoss} converges in 6h 52min, a greater than \textbf{13.5$\times$ speedup} over the 92h 47min required by the \textit{Set2Set} baseline. This shows that \textit{EERLoss} achieves its state-of-the-art accuracy (Table~\ref{tab:main_ablation}) through a more efficient training objective, with a training cost even better than the simple ArcFace and CosFace baselines.

This efficiency of the proposed algorithm is remarkable. Alg.~\ref{alg:ALPHABETA} reveals that the loss computation is embarrassingly parallel. This design avoids the iterative or complex sample-mining bottlenecks of other objectives, allowing it to scale efficiently on modern GPU hardware.

\vspace{4mm}

\section{Theoretical Grounding}
\label{sec:theo}
This section provides a theoretical justification for the performance gains observed with the area formulation of EERLoss relative to state-of-the-art alternative functions. Although angular losses are extensively adopted and optimized for physiological biometrics, they demonstrate suboptimal performance in comparison to EERLoss when applied to behavioral modalities such as keystroke dynamics. The underlying reasons for this phenomenon require further clarification.

A fundamental distinction between physiological and behavioral biometrics lies in their relative levels of signal stability. Physiological traits, such as faces or fingerprints, are anchored to permanent anatomical structures and thus exhibit significantly lower intra-subject variability than behavioral modalities. In the physiological domain, noise is typically extrinsic, arising from environmental interference or sensor corruption of an otherwise constant biological trait. Conversely, behavioral biometrics demonstrate inherent variability. In contrast to physiological modalities, behavioral profiles lack a fixed representation. In this case, intra-subject variability is not just a measurement artifact but an inherent property of the signal itself. While a face remains a definitive reference for a subject, there is no such definitive keystroke dynamics profile. Consequently, while physiological biometrics are characterized by well-behaved, predictable distributions, behavioral biometrics are inherently stochastic and non-ideal.

Angular loss functions impose four a priori assumptions on the structure of the embedding space: one topological and three geometrical. Topologically, they assume that embedding clusters lie on the surface of a hypersphere; that is, the manifold is closed and every geodesic eventually wraps back on itself. Geometrically, they assume that clusters have comparable radii, that they are approximately equidistant, and that inter-cluster overlap is negligible, allowing a separating margin. Like the physiological signals they encode, the resulting embedding clusters are well-behaved: compact, separable, and geometrically regular.

\begin{figure*}[t]
    \centering
    \includegraphics[width=\linewidth]{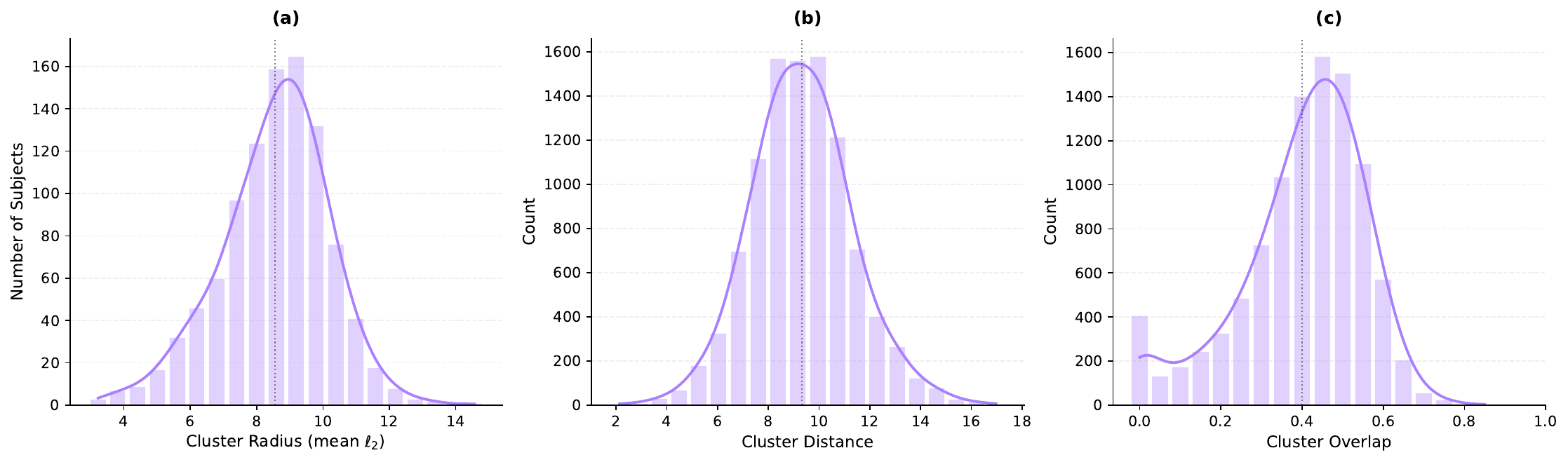}
    \vspace{-2mm}
    \caption{Empirical violation of geometric priors. We illustrate the non-ideal nature of keystroke embeddings through the distribution of: (a) cluster radii, showing high variance; (b) inter-cluster distances, showing non-equidistance; and (c) cluster overlap, which is non-negligible for the majority of subjects.}
    \label{fig:geometricViolations}
\end{figure*}

However, when applied to behavioral biometrics, these a priori assumptions become a liability. When the underlying dimensions of the space are not closed, projecting embeddings onto a hypersphere hinders classification by imposing false topological regularity. When inter-cluster overlap is significant, angular loss functions expend capacity attempting to enforce separation that the data cannot support; intrinsically inseparable clusters function as hard triplets, which are well-documented to destabilize training in metric learning frameworks. Similarly, constraining clusters to comparable radii distorts the embedding space when intra-subject variability profiles differ substantially across subjects. \textit{EERLoss}, by contrast, imposes no such priors and is free to learn the true topology and geometry of the embedding space from data.

We now substantiate the claim that keystroke dynamics profiles, across sufficiently large user populations, violate at least the three geometrical a priori assumptions imposed by angular loss functions: comparable cluster radii, approximate equidistance, and negligible inter-cluster overlap.

The geometric constraints are easier to disprove than the topological one. Figure \ref{fig:geometricViolations} (a) shows the distribution of per-subject cluster radii. Rather than concentrating around a common value, the distribution is broad, with radii spanning from 3 to 13, a fourfold range, indicating that cluster radii are far from uniform. Similarly, Figure \ref{fig:geometricViolations} (b) shows a broad distribution of distances to the nearest cluster center, disproving approximate equidistance. Finally, Figure 
\ref{fig:geometricViolations} (c) shows the distribution of cluster overlap between each 
cluster and its ten nearest neighbours. For clusters $i$ and $j$, overlap is defined as
\begin{equation}
    o_{ij} = \max\left(0, 1 - \frac{d_{ij}}{r_i + r_j}\right)
\end{equation}
where $d_{ij}$ is the Euclidean distance between centroids 
and $r_i, r_j$ are the respective cluster radii. Only a minority of nearest-neighbour cluster pairs are disjoint (leftmost bar, label 0.0, subfigure c); most exhibit overlaps averaging around 50\% and extending well beyond, directly violating the assumption of negligible inter-cluster overlap.

\begin{figure}[ht]
    \centering
    \includegraphics[width=\linewidth]{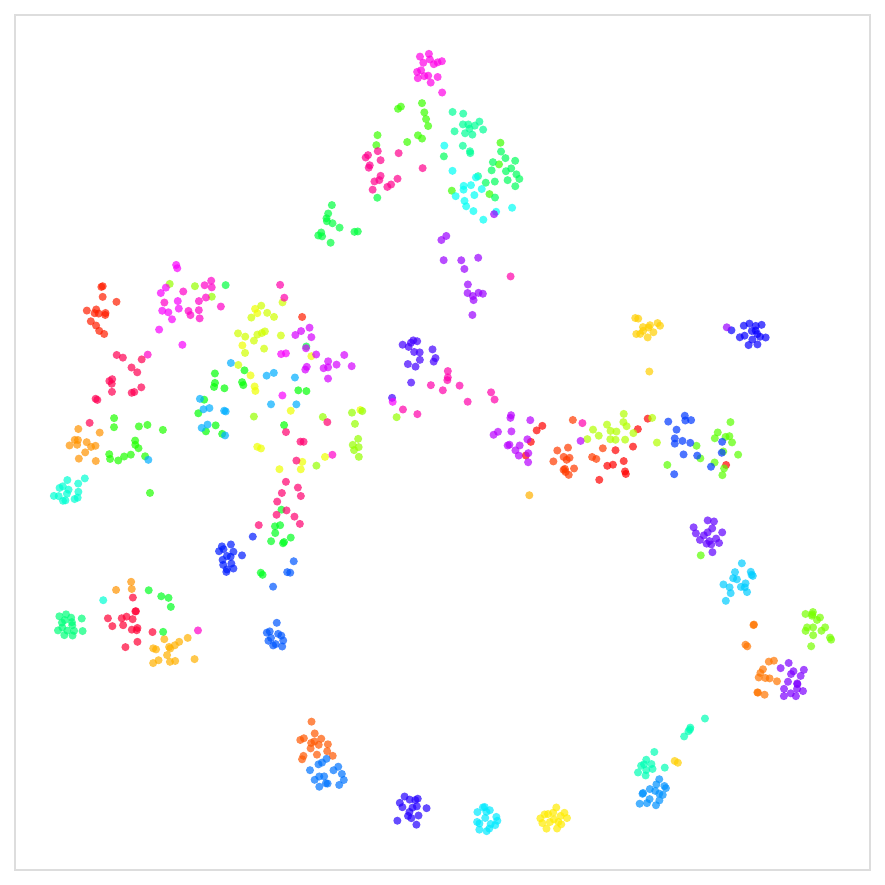}
    \vspace{-8mm}
    \caption{Distribution of subject clusters in the embedding space on the 1,000-user test set (EERLoss), dimensionality reduced via t-SNE. Rather than concentrating on a closed surface, clusters distribute along open, irregular manifolds.}
    \label{fig:clusters}
\end{figure}

\begin{figure}[ht]
    \centering
    \includegraphics[width=\linewidth]{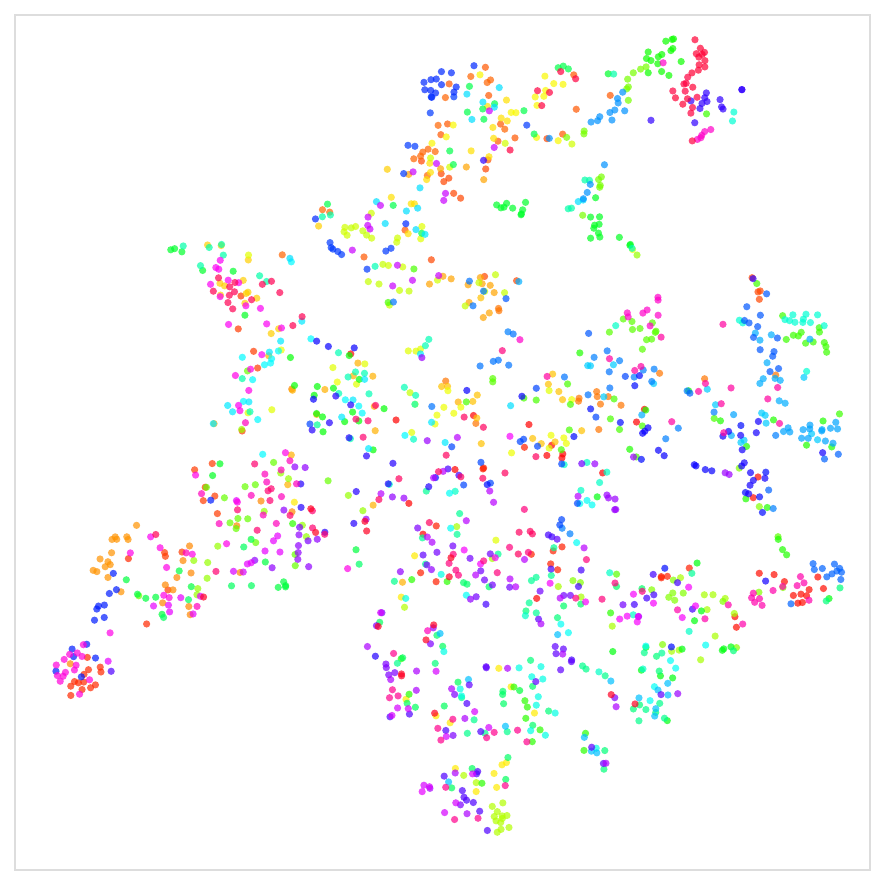}
    \vspace{-8mm}
    \caption{Distribution of subject clusters in the embedding space on the 1,000-user test set (ArcFace), dimensionality reduced via t-SNE. ArcFace enforces a more uniform use of the embedding space, in alignment with its a priori assumptions, at the cost of greater inter-cluster overlap compared to EERLoss.}
    \label{fig:clustersArcFace}
\end{figure}

Figure \ref{fig:clusters} shows a random selection of 50 clusters in the embedding space, reduced via t-SNE. The figure illustrates the violation of geometric assumptions at a glance: clusters exhibit significant mutual overlap and broad variation in radii and inter-cluster distances. Notably, the figure also refutes the topological assumption. Although dimensionality reduction via t-SNE introduces known local geometric distortions, the macroscopic open-manifold arrangement observed here remains fundamentally incompatible with a closed, hyperspherical prior. Instead, the clusters are arranged as elongated, irregular manifolds emanating from a common origin, consistent with an open rather than closed topology.

Figure \ref{fig:clustersArcFace} shows 50 clusters under ArcFace training, reduced via t-SNE for direct comparison. Here, the visual projection reflects ArcFace's a priori assumptions: the space is utilized more uniformly, suggesting a tendency towards comparable cluster radii and equidistant centroids. However, the trade-off of imposing this rigid geometry is visible in the boundaries between clusters, which are noticeably more porous than in Figure \ref{fig:clusters}, exhibiting greater overlap between neighbouring subjects. The resulting closed, radial arrangement stands in clear topological contrast to the open manifolds naturally learned by EERLoss.

The superior performance of \textit{EERLoss} in behavioral biometric tasks, and in keystroke dynamics in particular, can now be explained as stemming from its ability to adapt to the underlying geometry of the embedding space, rather than expending capacity enforcing a rigid geometric structure that the data do not support. Conversely, on theoretical grounds, angular loss functions should outperform \textit{EERLoss} in well-behaved classification tasks such as those arising in physiological biometrics, where their a priori assumptions align with the intrinsic structure of the data.

\subsection{Case Study on Face Recognition Under Data Scarcity}

The substantial improvements reported in our results indicate that \textit{EERLoss} effectively exploits the irregular nature of behavioral biometrics. To explore the boundaries of this performance and test how our approach generalizes to more structured data, we conducted a preliminary evaluation on face recognition, one of the main physiological modalities, characterized by more stable geometric profiles. We trained an IResNet50 on subsets of CASIA-WebFace\cite{yi2014casia} and evaluated on LFW \cite{huang2008labeled}, deliberately limiting the training samples per identity ($N_{img}$) to simulate different levels of data scarcity.

Table \ref{tab:face_rec} presents the results of this small-scale comparison. As expected, in a well-behaved modality like face recognition with abundant data ($N_{img}=16$), ArcFace \cite{deng2019arcface} maintains its superiority, as its hyperspherical priors align well with the densely sampled embedding space. However, as we move towards more restrictive regimes, the gap narrows. Notably, in the most extreme case of data scarcity ($N_{img}=4$), \textit{EERLoss} manages to outperform ArcFace, reducing the EER from 13.92\% to 11.75\%. 

This suggests that while angular losses remain the optimal choice for physiological biometrics under ideal sampling conditions, \textit{EERLoss} offers a competitive alternative when global geometric proxies (like class centers) cannot be robustly estimated. More importantly, it confirms that the strength of \textit{EERLoss} lies in its ability to operate without rigid assumptions, making it particularly valuable for biometric traits where intra-subject variability is high and training samples are limited.

\begin{table}[t]
\caption{Preliminary evaluation on Face Verification (LFW) under different data regimes to assess generalization limits.}
\label{tab:face_rec}
\centering
\small
\begin{tabular}{@{}lcccc@{}}
\toprule
\textbf{Setup ($N_{img}$)} & \textbf{Loss} & \textbf{Acc (\%)} & \textbf{EER (\%)} & \textbf{Gain (EER)} \\ \midrule
\multirow{2}{*}{\textbf{16 imgs/ID}} & ArcFace & \textbf{96.76} & \textbf{3.38} & -- \\
                             & EERLoss & 95.97 & 4.08 & -0.7\% \\ \midrule
\multirow{2}{*}{\textbf{8 imgs/ID}}  & ArcFace & 92.08 & 7.55 & -- \\
                             & EERLoss & \textbf{93.28} & \textbf{6.80} & \textbf{+0.75\%} \\ \midrule
\multirow{2}{*}{\textbf{4 imgs/ID}}  & ArcFace & 86.43 & 13.92 & -- \\
                             & EERLoss & \textbf{87.96} & \textbf{11.75} & \textbf{+2.17\%} \\ \bottomrule
\end{tabular}
\end{table}

\section{Conclusion}
\label{sec:conclusoin}
\vspace{-1mm}
In this paper we introduce \textit{EERLoss}, a novel loss function that enables direct end-to-end optimization of the target metric. A rigorous evaluation was conducted on the large-scale KVC-onGoing benchmark. The empirical results are conclusive: \textit{EERLoss} significantly outperforms the SoTA, achieving a relative EER reduction of up to 30.7\%. Furthermore, the proposed method converges substantially faster compared to other losses, reducing the SoTA's overall training cost from 92 hours and 47 minutes to only 6 hours and 52 minutes in the final experiment. These results strongly validate this approach and suggest that direct metric optimization, via \textit{EERLoss}, is a superior paradigm for training deep biometric verification models. The scope of this study’s empirical validation was focused on the keystroke dynamics modality. This was a deliberate methodological choice, using the KVC-onGoing benchmark as a rigorous, large-scale test known for its high intra-class and low inter-class variability. While there is already strong evidence of the method’s efficacy, future work will investigate the generalization of \textit{EERLoss} to other verification tasks. Additionally, the underlying subdifferentiable approximation framework itself could be adapted to optimize other non-differentiable evaluation metrics present in computer vision and beyond.

\section{Acknowledgment}
This project has been supported by Cátedra ENIA UAM-VERIDAS en IA Responsable (NextGenerationEU PRTR TSI100927-2023-2) and TRUST-ID (PID2025-173396OB-I00 MICIU/AEI and the EU). Robledo-Moreno is supported by a FPI Fellowship (FPI-UAM-2025).

\bibliographystyle{IEEEtran}
\bibliography{main}

@article{chechik2010large,
  title={{Large Scale Online Learning of Image Similarity Through Ranking.}},
  author={Chechik, Gal and Sharma, Varun and Shalit, Uri and others},
  journal={Journal of Machine Learning Research},
  volume={11},
  number={3},
  year={2010}
}

@article{morales2022setmargin,
  title={{SetMargin Loss Applied to Deep Keystroke Biometrics with Circle Packing Interpretation}},
  author={Morales, Aythami and Fierrez, Julian and Acien, Alejandro and others},
  journal={Pattern Recognition},
  year={2022},
  publisher={Elsevier}
}

@article{gonzalez2024type2branch,
  title={{Type2Branch: Keystroke Biometrics Based on a Dual-Branch Architecture With Attention Mechanisms and Set2set Loss}},
  author={Gonzalez, Nahuel and Stragapede, Giuseppe and Vera-Rodriguez, Ruben and others},
  journal={IEEE Transactions on Information Forensics and Security},
  year={2025}
}

@article{terven2025comprehensive,
  title={{A Comprehensive Survey of Loss Functions and Metrics in Deep Learning}},
  author={Terven, Juan and Cordova-Esparza, Diana-Margarita and Romero-Gonzalez, Julio-Alejandro and others},
  journal={Artificial Intelligence Review},
  year={2025},
  publisher={Springer}
}

@inproceedings{deng2019arcface,
  title={{Arcface: Additive Angular Margin Loss for Deep Face Recognition}},
  author={Deng, Jiankang and Guo, Jia and Xue, Niannan and others},
  booktitle={In Proc. of the IEEE/CVF Conference on Computer Vision and Pattern Recognition},
  year={2019}
}

@inproceedings{wang2018cosface,
  title={{Cosface: Large Margin Cosine Loss for Deep Face Recognition}},
  author={Wang, Hao and Wang, Yitong and Zhou, Zheng and others},
  booktitle={In Proc. of the IEEE Conference on Computer Vision and Pattern Recognition},
  year={2018}
}

@article{schultz2003triplet,
  title={{Learning a Distance Metric from Relative Comparisons}},
  author={Schultz, Matthew and Joachims, Thorsten},
  journal={Advances in neural information processing systems},
  year={2003}
}

@inproceedings{chopra2005contrastive,
  title={{Learning a Similarity Metric Discriminatively, with Application to Face Verification}},
  author={Chopra, Sumit and Hadsell, Raia and LeCun, Yann},
  booktitle={In Proc. of the IEEE Conference on Computer Vision and Pattern Recognition},
  year={2005},
}

@article{shadman2025keystrokesurvey,
  title={{Keystroke Dynamics: Concepts, Techniques, and Applications}},
  author={Shadman, Rashik and Wahab, Ahmed Anu and Manno, Michael and others},
  journal={ACM Computing Surveys},
  year={2025},
  publisher={ACM New York, NY}
}

@article{stragapede2025kvc,
  title={{KVC-onGoing: Keystroke Verification Challenge}},
  author={Stragapede, Giuseppe and Vera-Rodriguez, Ruben and Tolosana, Ruben and others},
  journal={Pattern Recognition},
  year={2025},
  publisher={Elsevier}
}

@inproceedings{dhakal2018aaltodesktop,
  title={{Observations on Typing from 136 Million Keystrokes}},
  author={Dhakal, Vivek and Feit, Anna Maria and Kristensson, Per Ola and others},
  booktitle={In Proc. of the 2018 CHI Conference on Human Factors in Computing Systems},
  year={2018}
}

@inproceedings{palin2019aaltomobile,
  title={{How do People Type on Mobile Devices? Observations from a Study with 37,000 Volunteers}},
  author={Palin, Kseniia and Feit, Anna Maria and Kim, Sunjun and others},
  booktitle={In Proc. of the 21st International Conference on Human-Computer Interaction with Mobile Devices and Services},
  year={2019}
}

@INPROCEEDINGS{2009greyc,
  author={Giot, Romain and El-Abed, Mohamad and Rosenberger, Christophe},
  booktitle={In Proc. IEEE International Conference on Biometrics: Theory, Applications, and Systems}, 
  title={{GREYC Keystroke: A Benchmark for Keystroke Dynamics Biometric Systems}}, 
  year={2009}}

@article{2014rhu,
author = {El-Abed, Mohamad and Dafer, Mostafa and El Khayat, Ramzi},
year = {2014},
title = {{RHU Keystroke: A Mobile-Based Benchmark for Keystroke Dynamics Systems}},
journal = {Proceedings - International Carnahan Conference on Security Technology}}

@inproceedings{2017clarkson,
author = {Murphy, Christopher and Huang, Jiaju and Hou, Daqing and others},
title = {{Shared Dataset on Natural Human-Computer Interaction to Support Continuous Authentication Research}},
year = {2017},
booktitle = {In Proc. IEEE International Joint Conference on Biometrics}
}

@ARTICLE {BeCAPTCHA_BotDetectionHuMiDB,
author = {Alejandro Acien and Aythami Morales and Julian Fierrez and others},
journal = {Engineering Applications of Artificial Intelligence},
publisher = {Elsevier},
title = {{BeCAPTCHA: Behavioral Bot Detection using Touchscreen and Mobile Sensors benchmarked on HuMIdb}},
year = {2021}
}

@inproceedings{maiorana2025signature,
  author    = {Maiorana, Emanuele and others},
  title     = {{Signature Biometrics}},
  booktitle = {Encyclopedia of Cryptography, Security and Privacy},
  year      = {2025},
  publisher = {Springer}
}

@ARTICLE{delgado_gait,
author = {Paula Delgado-Santos and Ruben Tolosana and others},
journal = {Engineering Applications of Artificial Intelligence},
title = {{{M-GaitFormer}: Mobile Biometric Gait Verification Using Transformers}},
year = {2023},
}

@ARTICLE{tolo_deep,
  author={Ruben Tolosana and Ruben Vera-Rodriguez and Julian Fierrez and others},
  journal={IEEE Transactions on Biometrics, Behavior, and Identity Science}, 
  title={{DeepSign: Deep On-Line Signature Verification}}, 
  year={2021},
}

@article{jain2006biometrics,
  title={{Biometrics: A Tool for Information Security}},
  author={Jain, Anil K and Ross, Arun and Pankanti, Sharath},
  journal={IEEE Transactions on information forensics and security},
  year={2006}
}

@inproceedings{kim2022adaface,
  title={{Adaface: Quality adaptive margin for face recognition}},
  author={Kim, Minchul and Jain, Anil K and Liu, Xiaoming},
  booktitle={Proceedings of the IEEE/CVF conference on computer vision and pattern recognition},
  year={2022}
}

@article{yi2014casia,
  title={{Learning Face Representation from Scratch}},
  author={Yi, Dong and Lei, Zhen and Liao, Shengcai and Li, Stan Z},
  journal={arXiv preprint arXiv:1411.7923},
  year={2014}
}

@inproceedings{huang2008labeled,
  title={{Labeled Faces in the Wild: A Database for Studying Face Recognition in Unconstrained Environments}},
  author={Huang, Gary B and Mattar, Marwan and Berg, Tamara and Learned-Miller, Eric},
  booktitle={Workshop on faces in'Real-Life'Images: detection, alignment, and recognition},
  year={2008}
}

\section{Biography Section}
\vspace{-10mm}

\begin{IEEEbiography}[{\includegraphics[width=1in,height=1.25in,clip,keepaspectratio]{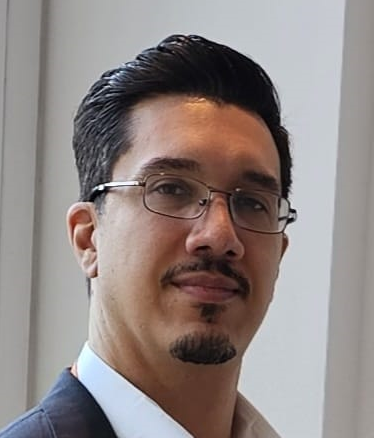}}]{Nahuel González} received his Ph.D. degree in Computing Science from Universidad Nacional de La Plata (UNLP), Argentina, in 2022. His Ph.D. thesis was awarded the Raúl Gallard Prize, handed by the Network of Argentinian Universities of Computing Science (RedUNCI), the following year. He has been affiliated with the Laboratorio de Sistemas de Información Avanzados (LSIA) of the University of Buenos Aires (UBA) since 2013. His main research interests are behavioral biometrics and time series prediction/classification using deep learning. He is a member of the editorial board of Data in Brief, Elsevier, since 2024.
\end{IEEEbiography}

\vspace{-10mm}

\begin{IEEEbiography}[{\includegraphics[width=1in,height=1.25in,clip,keepaspectratio]{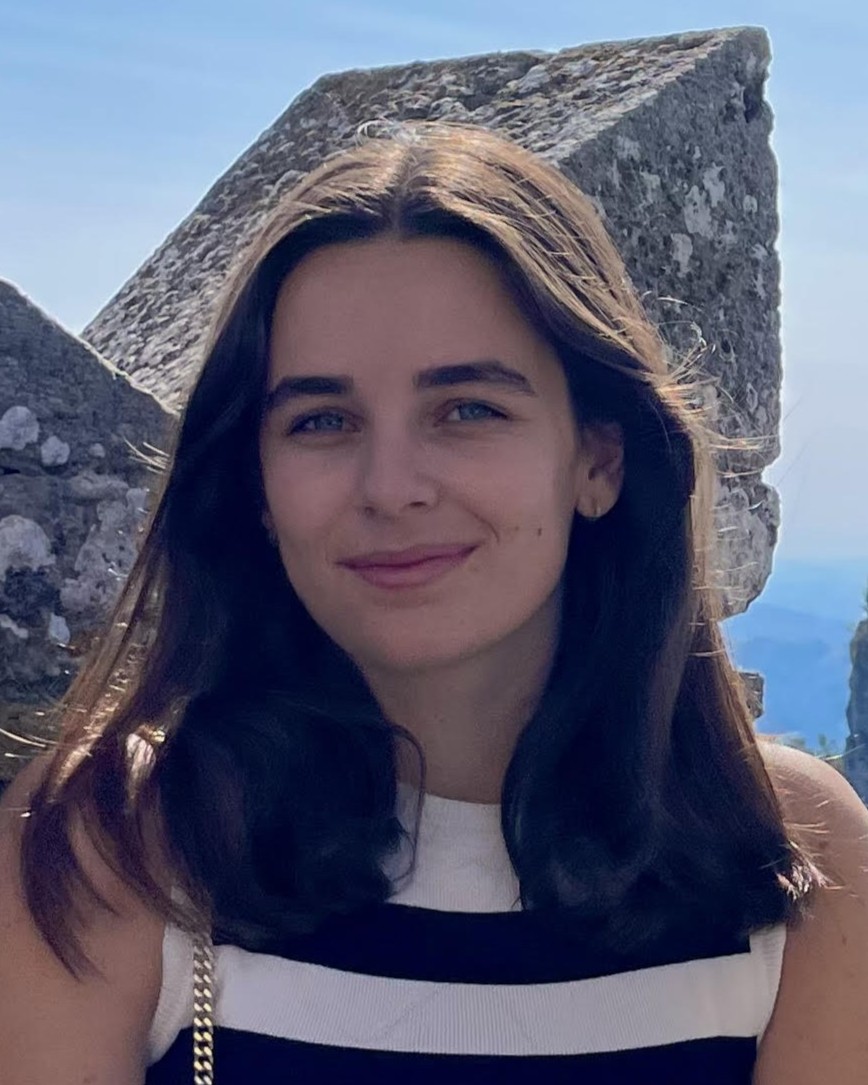}}]{Marta Robledo-Moreno} received her B.Sc. and M.Sc. degrees in Telecommunication Engineering from Universidad Autonoma de Madrid (Spain) in 2022 and 2024, respectively. The beginning of her research career was at the Biometric System Laboratory of the University of Bologna (Italy). Subsequently, she joined the BiometricsAI Lab (UAM), where she is currently collaborating as an assistant researcher pursuing a Ph.D. degree. Her research interests are mainly focused on signal processing and deep learning for behavioral biometrics.
\end{IEEEbiography}
\vspace{-10mm}

\vfill
\pagebreak

\begin{IEEEbiography}[{\includegraphics[width=1in,height=1.25in,clip,keepaspectratio]{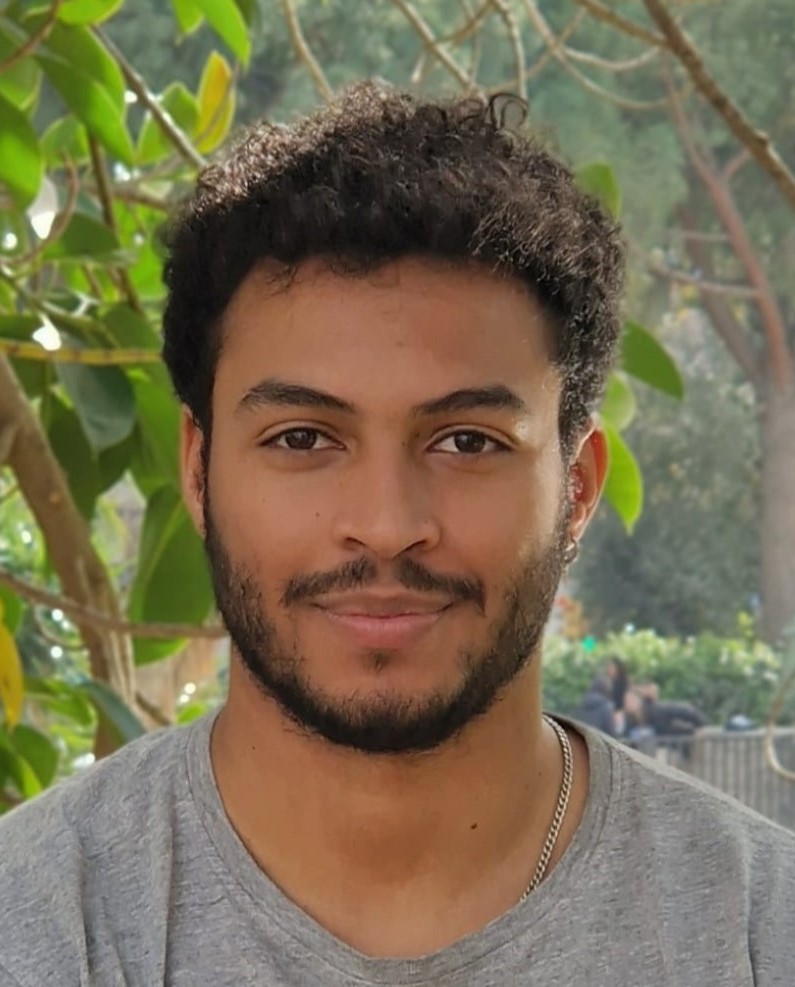}}]{Ivan Deandres-Tame} received the B.Sc. in Computer Science and Engineering in 2021 and the M.Sc. in Deep learning for Image, Audio and Video processing in 2022 from the Universidad Autonoma de Madrid. In May 2023, he joined the BiometricsAI Lab (UAM), where he is currently collaborating as an assistant researcher pursuing a Ph.D. degree. The research activities he is currently working are focused in Person recognition in complex scenarios and the use of synthetic data to mitigate capture challenges.
\end{IEEEbiography}
\vspace{-10mm}

\begin{IEEEbiography}[{\includegraphics[width=1in,height=1.25in,clip,keepaspectratio]{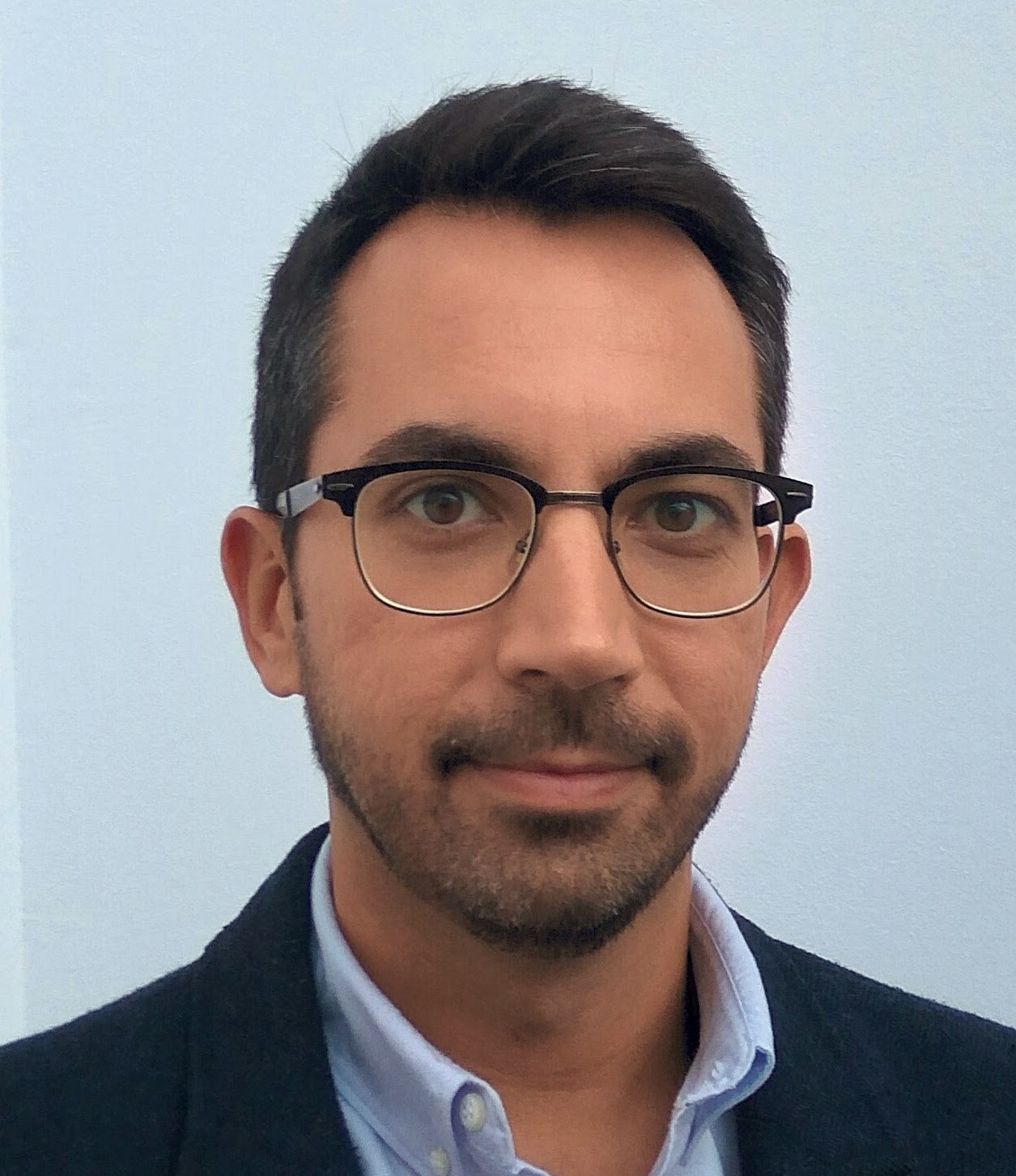}}]{Ruben Vera-Rodriguez} received his PhD degree in electrical and electronic engineering from Swansea University, U.K., in 2010. Since then, he has been affiliated with the BiometricsAI Lab, Universidad Autonoma de Madrid, Spain, where he is currently an Associate Professor since 2018. His research interests include signal and image processing, pattern recognition, HCI and biometrics, with emphasis on signature, face, gait verification, mobile biometrics and forensic applications of biometrics. He is actively involved in several national and European projects focused on biometrics. He has been awarded recently with a Medal from the Spanish Royal Academy of Engineering for his research contributions. He is member of ELLIS Society.
\end{IEEEbiography}

\vspace{-10mm}

\begin{IEEEbiography}[{\includegraphics[width=1in,height=1.25in,clip,keepaspectratio]{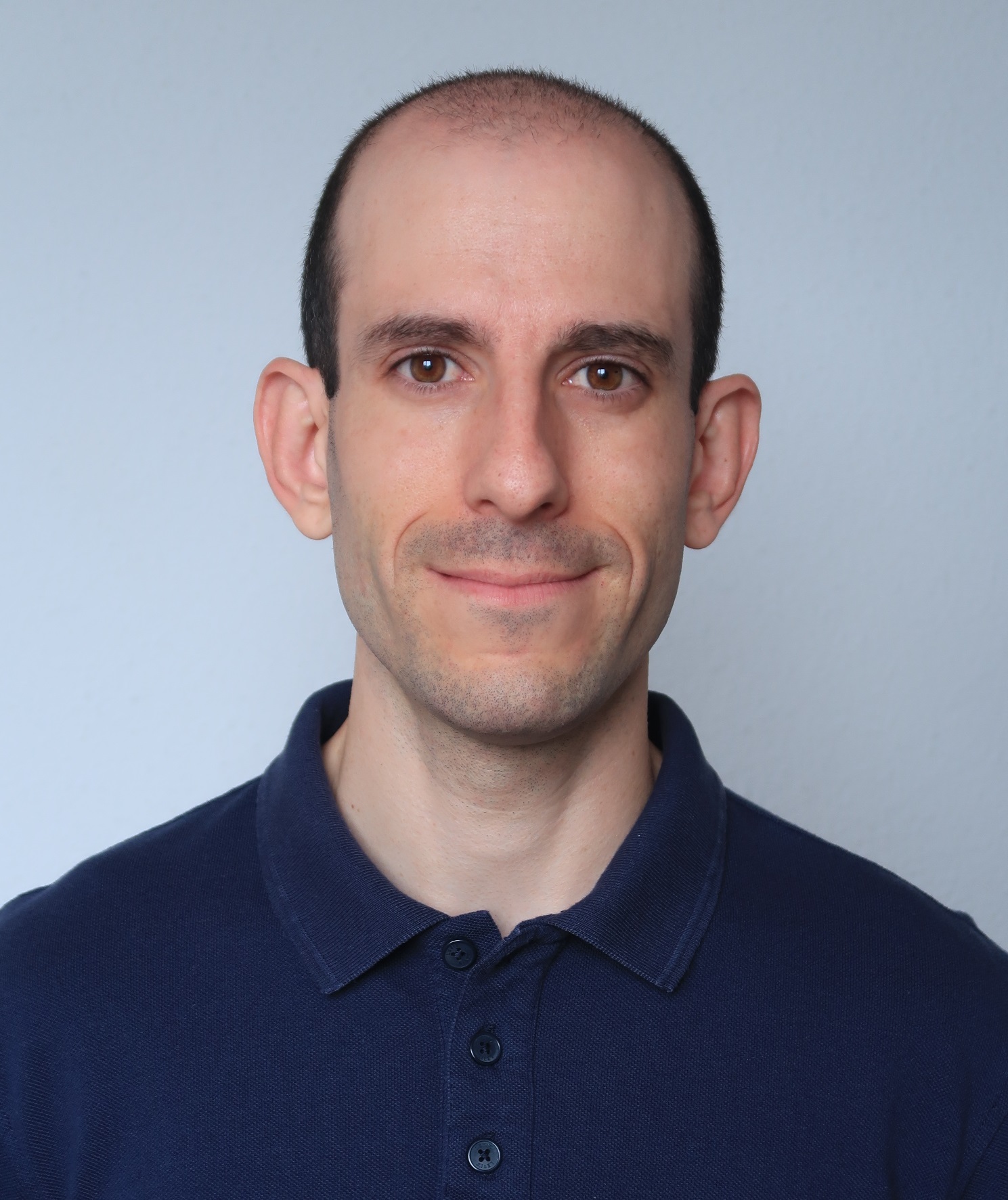}}]{Ruben Tolosana} received the M.Sc. degree
in Telecommunication Engineering, and the Ph.D. degree in Computer and Telecommunication Engineering, from Universidad Autonoma de Madrid, in 2014 and 2019, respectively. In 2014, he joined the BiometricsAI Lab at the Universidad Autonoma de Madrid, where he is currently an Assistant Professor. He is a member of the ELLIS Society, the Technical Area Committee of EURASIP, and the Editorial Board of the IEEE Biometrics Council Newsletter. His research interests are mainly focused on signal and image processing, pattern recognition, and machine learning, particularly in the areas of DeepFakes, Human-Computer Interaction, Biometrics, and Health. Dr. Tolosana has also received several awards such as the European Biometrics Industry Award (2018) from the European Association for Biometrics (EAB) and the Best Ph.D. Thesis Award in 2019-2022 from the Spanish Association for Pattern Recognition and Image Analysis (AERFAI).
\end{IEEEbiography}

\vfill

\end{document}